\pdfoutput=1

\documentclass[11pt]{article}
\usepackage{acl}

\usepackage{times}
\usepackage{latexsym}
\usepackage{graphicx}
\usepackage[utf8]{inputenc}
\usepackage{times}
\usepackage{soul}
\usepackage{url}
\usepackage{amsmath}
\usepackage{amsthm}
\usepackage{booktabs}
\usepackage{algorithm}
\usepackage{algorithmicx}
\usepackage{algpseudocode}
\usepackage{microtype}
\usepackage{hyperref}
\usepackage{caption}
\usepackage{arydshln}
\usepackage{pgfplots} 
\usepackage{verbatim}
\usepackage{bbm}
\usepackage{amssymb}
\usepackage{multirow}
\usepackage{color}
\usepackage[T1]{fontenc}
\usepackage{xcolor}
\usepackage[utf8]{inputenc}
\usepackage{microtype}

\title{Type-Driven Multi-Turn Corrections for Grammatical Error Correction}

\author{Shaopeng Lai$^{1}$\thanks{\quad Work is done during internship at Tencent Cloud Xiaowei}, Qingyu Zhou$^2$, Jiali Zeng$^2$, Zhongli Li$^2$, \\\textbf{Chao Li}$^2$, \textbf{Yunbo Cao}$^2$,  \textbf{Jinsong Su}$^{1,3}$\thanks{\quad Corresponding author}\\
$^{1}$School of Informatics, Xiamen University, China  \\$^{2}$Tencent Cloud Xiaowei, China\\
$^{3}$Key Laboratory of Digital Protection and Intelligent Processing of Intangible \\ Cultural Heritage of Fujian and Taiwan, Ministry of Culture and Tourism, China \\
splai@stu.xmu.edu.cn, \{qingyuzhou, lemonzeng, neutrali, diegoli, yunbocao\}@tencent.com, \\
jssu@xmu.edu.cn}

\begin{document}
\maketitle
\begin{abstract}
Grammatical Error Correction (GEC) aims to automatically detect and correct grammatical errors.
In this aspect, dominant models are trained by one-iteration learning while performing multiple iterations of corrections during inference. 
Previous studies mainly focus on the data augmentation approach to combat the exposure bias, which suffers from two drawbacks. 
First, they simply mix additionally-constructed training instances and original ones to train models, which fails to help models be explicitly aware of the procedure of gradual corrections. Second, they ignore the interdependence between different types of corrections.
In this paper, we propose a \textit{Type-Driven Multi-Turn Corrections} approach for GEC.
Using this approach, from each training instance, we additionally construct multiple training instances, each of which involves the correction of a specific type of errors. Then, we use these additionally-constructed training instances and the original one to train the model in turn.
By doing so, our model is trained to not only correct errors progressively, but also exploit the interdependence between different types of errors for better performance.
Experimental results and in-depth analysis show that our approach significantly benefits the model training.
Particularly, our enhanced model achieves state-of-the-art single-model performance on English GEC benchmarks. We release our code at \url{https://github.com/DeepLearnXMU/TMTC}.

\end{abstract}

\section{Introduction}\label{section:introduction}

Grammatical Error Correction (GEC) aims at automatically detecting and correcting grammatical (and other related) errors in a text. 
It attracts much attention due to its practical applications in writing assistant \cite{Napoles_EACL2017,Ghufron_LC18}, speech recognition systems \cite{Karat_CHI99,Wang_INTERSPEECH20, Kubis_arxiv20O} etc. 
Inspired by the success of neural machine translation (NMT), some models adopt the same paradigm, namely NMT-based models. They have been quite successful, especially with data augmentation approach \cite{Boyd_EMNLP18, Ge_ACL18, Xu_ACL19,Grundkiewicz_ACL19,Wang_EMNLP20,Takahashi_ACL20}. However, these models have been blamed for their inefficiency during inference \cite{Chen_EMNLP20,Sun_ACL2021}. 
To tackle this issue, many researchers resort to the sequence-to-label (Seq2Label) formulation, achieving comparable or better performance with efficiency \cite{Malmi_emnlp2019,  awasthi_emnlp2019, stahlberg_emnlp2020, omelianchuk_gector_ACL20}.

Despite their success,
both NMT-based and Seq2Label models 
are trained by one-iteration learning, 
while correcting errors for multiple iterations during inference. As a consequence, they suffer from exposure bias and exhibit performance degrade
\cite{Ge_ACL18,Lichtarge_NAACL19,Zhao_AAAI20,Parnow_aclf21}.
To deal with this issue, 
\citet{Ge_ACL18} 
propose to generate fluency-boost pseudo instances as additional training data.
Besides, \citet{Parnow_aclf21} 
dynamically augment training data by introducing the predicted sentences with high error probabilities.

However, the above-mentioned approaches construct pseudo data based on a GEC model or an error-generation model, which extremely depends on the performance of these models. Thus, the error distribution of pseudo data is biased and lacks diversity and practicality. Moreover, they simply mix original and pseudo data to train models, which are unable to learn correcting errors progressively. 
Furthermore, they ignore the interdependence between different types of errors, which intuitively plays an important role on GEC. Taking Table \ref{tab:case} as example, correcting ``\textit{little}'' with ``\textit{few}'' or ``\textit{job}'' with ``\textit{jobs}'' first can help the other error be better corrected. Therefore, we believe that how to construct and exploit pseudo data with editing-action corrections for GEC is still a problem worthy of in-depth study.

\begin{table}[t] \footnotesize
    \centering
    \begin{tabular}{l}
    \toprule
    Erroneous Sentence: \textit{In my country there are \textbf{\textcolor{cyan}{little} \textcolor{magenta}{job}}  } \\ 
    \textit{because the economy is very bad .}\\
    \midrule
    Reference Sentence: \textit{In my country there are \textbf{\textcolor{cyan}{few} \textcolor{magenta}{jobs}} } \\ 
    \textit{because the economy is very bad .}\\
    \bottomrule
    \end{tabular}
    \caption{An example for the interdependence between corrections. Please note that whichever error is corrected first, the other error can be corrected more easily.}
    \label{tab:case}
    \vspace{-5pt}
\end{table}

In this paper,
we first conduct quantitative experiments to investigate the performance improvements of GEC model with providing different types of error corrections. Experimental results show that corrections of appending or replacing words first indeed benefit the corrections of other errors.
Furthermore, we propose a {\bf T}ype-Driven {\bf M}ulti-{\bf T}urn {\bf C}orrections (TMTC) approach for GEC. Concretely, by correcting a certain type of errors with others unchanged, we construct an intermediate sentence for each training instance and pair it with its raw erroneous sentence and reference sentence respectively, forming two additional training instances.
During the model training,
using the former instance, 
we firstly guide the model to learn correcting the corresponding type of errors.
Then, 
using the latter instance, 
we teach the model to correct other types of errors with the help of previous corrections.
Overall, contributions of our work are three-fold:
\begin{itemize}
    \vspace{-5pt}
    \item Through quantitative experiments, we investigate the interdependence between different types of corrections, with the finding that corrections of appending or replacing words significantly benefit correcting other errors.
    \vspace{-5pt}
    \item We propose a TMTC approach for GEC. To the best of our knowledge, our work is the first attempt to explore the interdependence between different types of errors for GEC.
    \vspace{-5pt}
    \item We conduct experiments and in-depth analysis to investigate the effectiveness of our proposed approach. Experimental results show that our enhanced model achieves the state-of-the-art performance.

\end{itemize}

\section{Related Work}\label{sec:related_work}

Generally, there are two categories of models in GEC: Transformer-dominant NMT-based models \cite{Boyd_EMNLP18, Ge_ACL18, Xu_ACL19, Grundkiewicz_ACL19, Wang_EMNLP20, Takahashi_ACL20} and GECToR-leading Seq2Label models \cite{Malmi_emnlp2019,  awasthi_emnlp2019, stahlberg_emnlp2020, omelianchuk_gector_ACL20}. The former models consider GEC as a machine translation task, where the model is fed with the erroneous sentence and then output the corrected sentence token by token. By comparison, Seq2Label models are able to correct grammatical errors more efficiently and even better. Among them, the GECToR models \cite{omelianchuk_gector_ACL20} obtain remarkable performance. Typically, they adopt a pre-trained language model as the encoder to learn word-level representations and utilize a softmax-based classifier to predict designed editing-action labels. 

\begin{figure*}[t]
  \centering
  \includegraphics[width=1\linewidth]{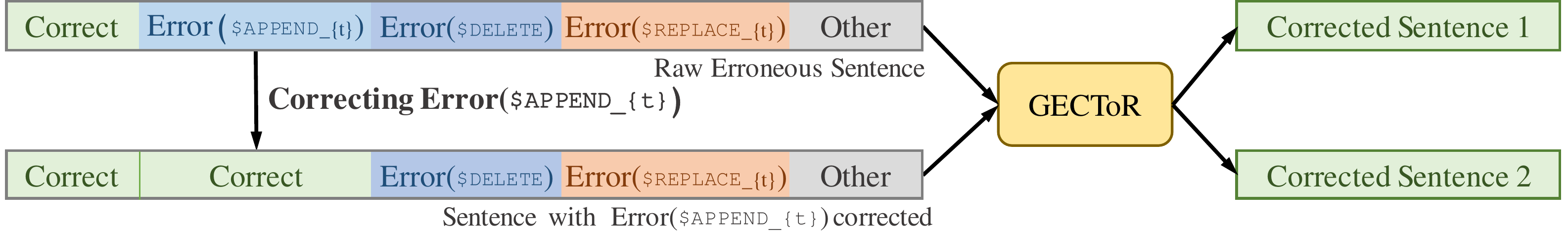}
  \vspace{-15pt}
  \caption{The procedure of our quantitative experiments. Each sentence is composed of five parts as illustrated, where Error(\texttt{ACTION} \textit{label}) denote the erroneous words that can be corrected via corresponding editing-action label. We only correct one type of errors and compare the prediction results of other types of errors.}
  \label{fig:control_expriment}

\end{figure*}

Since GEC models may fail to completely correct a sentence through just one-iteration inference, some researchers resort to data augmentation that has been widely used in other NLP studies \cite{song_acl2020,song_aaai2020}. For instance, \citet{Ge_ACL18} propose to let the GEC model infer iteratively and design a fluency boost learning approach. Specifically, they establish new erroneous-reference sentence pairs by pairing predicted less fluent sentences with their reference sentences during training.
Likewise, to solve the mismatches between training and inference of Seq2Label models, \citet{Parnow_aclf21} apply a confidence-based method to construct additional training data by pairing low-confidence sentences with reference sentences. 
Note that these two methods also involve constructing pseudo data using sentences with partial errors.
However, ours is still different from them in two aspects.
First, 
these two methods simply mix their pseudo data with original data to still train models in a one-iteration learning manner. By contrast, we decompose the one-iteration corrections into multiple turns, so as to make the model aware of gradual corrections.
Second, 
these two methods ignore the interdependence between different types of errors, which is exploited by our proposed approach to enhance the model.

\section{Background}

In this work, we choose GECToR \cite{omelianchuk_gector_ACL20} as our basic GEC model due to its efficiency and competitive performance.
Typically, it considers the GEC task as a sequence-to-label task, where the candidate editing-action labels mainly include \texttt{\$KEEP} (to keep the current word unchanged), \texttt{\$DELETE} (to delete the current word), \texttt{\$APPEND\_\{t\}} (to append the word $t$ after the current word), \texttt{\$REPLACE\_\{t\}} (to replace the current word with the word $t$) and some elaborate g-transformation labels \cite{omelianchuk_gector_ACL20} performing task-specific operations, such as \texttt{\$TRANSFORM\_CASE\_LOWER} and \texttt{\$TRANSFORM\_CASE\_CAPITAL} (to change the case of the current word).

On the whole, the GECToR model is composed of an encoder based on pre-trained language model and two linear classifiers: one for grammatical error detection (GED) and the other for GEC. The encoder reads the erroneous sentence $X_e$=$x_1,x_2,...,x_N$ and represent words with hidden states $\{h_i\}_{i=1}^N$, which are fed into classifiers to predict the binary label sequence $Y$=$y_1,y_2,...,y_N$ for GED and the editing-action label sequence $T$=$t_1,t_2,...,t_N$ for GEC, respectively. Formally, the losses of two classifiers can be formulated as
\vspace{-5pt}
\begin{equation} \label{eq:baseline_loss}
	\begin{aligned}
		L_d&=-\sum_{i=1}^{N}\text{log} p(y_i|X_e,\theta), \\ 
	\end{aligned}
\end{equation}
\vspace{-7pt}
\begin{equation} \label{eq:baseline_loss}
	\begin{aligned}
		L_c&=-\sum_{i=1}^{N}\text{log} p(t_i|X_e,\theta),\\
	\end{aligned}
\end{equation}where $\theta$ denotes model parameters. Usually, the GECToR model is trained to optimize the sum of two losses: $L$=$L_d$+$L_c$.

It is worth noting that the GECToR model is trained to correct all errors in a one-iteration manner, while correcting errors in a multiple-iteration way during inference (at most 5 iterations). 
\begin{table}[ht] \footnotesize
	\centering
	\setlength{\tabcolsep}{0.3mm}{
		
		\begin{tabular}{lccccc} 
			\toprule
			 \multicolumn{1}{c}{Dataset}  & &\#Instance & & Stage  \\
			\midrule
			PIE-synthetic \cite{awasthi_emnlp2019}	& & 9,000,000 & & \uppercase\expandafter{\romannumeral1} \\[0.2ex]
			Lang-8 \cite{tajiri_acl2012_lang8} 	& & 947,344 & & \uppercase\expandafter{\romannumeral2} \\[0.2ex]
			NUCLE \cite{dahlmeier_2013_NUCLE} & & 56,958 & & \uppercase\expandafter{\romannumeral2} \\[0.2ex]
			FCE \cite{Yannakoudakis_acl2011_FCE} & & 34,490 & & \uppercase\expandafter{\romannumeral2} \\[0.2ex]
			W\&I+LOCNESS \cite{bryant_2019_wilocness} & & 34,304 & & \uppercase\expandafter{\romannumeral2}, \uppercase\expandafter{\romannumeral3} \\
			\bottomrule
	\end{tabular}
	}
	\caption{GECToR is trained on PIE-synthetic dataset for pre-training at Stage \uppercase\expandafter{\romannumeral1}. Then, it is fine-tuned on Lang-8, NUCLE, FCE, W\&I+LOCNESS at Stage \uppercase\expandafter{\romannumeral2}. At Stage \uppercase\expandafter{\romannumeral3}, the final fine-tuning is conducted on W\&I+LOCNESS. }
	\label{tab:datasets}
\end{table}
Besides, there are three stages involved during the training of the GECToR model, as shown in Table \ref{tab:datasets}.

\begin{table*}[ht]\small
	\centering
	\setlength{\tabcolsep}{1.7mm}{
		\begin{tabular}{cccccccccc} 
			\toprule
			\multicolumn{1}{c}{\multirow{3}{*}{Dataset}}    &    &\multicolumn{8}{c}{RoBERTa} \\ 
			\cline{3-10}
			 & Evaluation &\multicolumn{4}{c}{BEA-2019 (dev)} &\multicolumn{4}{c}{CoNLL-2014 (test)} \\
			\cline{3-10}
			& & Num. & Prec. & Rec. & F$_1$ & Num. & Prec. & Rec. & F$_1$ \\
			\hline
		  \multicolumn{1}{c}{\multirow{3}{*}{Original Dataset}} &\texttt{\$APPEND\_\{t\}} & 2609 & 53.43 & 35.22 & 42.46 & 621 & 27.46 &23.35 &25.24 \\
		   &\texttt{\$DELETE} & 1403 & 56.04 & 23.81 & 33.42 & 1115 & 51.89 &18.48 &27.25 \\
		  {} &\texttt{\$REPLACE\_\{t\}} & 3495 & 50.87 & 23.32 & 31.98 & 1398 & 38.57 &18.45 &24.96 \\
			\hline
		  \multicolumn{1}{c}{\multirow{2}{*}{$D$(\texttt{APPEND})}} &\texttt{\$DELETE}    & 904 & 62.63 & 20.02 & 30.34 & 496 & 47.52 &13.51 &21.04 \\
		   &\texttt{\$REPLACE\_\{t\}}   & 2079 & 49.71 & 20.30 & 28.83 & 660 & 28.57 &11.21 &16.10 \\[0.2ex]
			\hdashline[2pt/2pt]	
			\multicolumn{1}{c}{\multirow{2}{*}{$D$(\texttt{APPEND\checkmark})}} &\texttt{\$DELETE}  & 904 & 68.84 & 26.88 & 38.66 (+8.32) & 496 & 59.06 &17.74 &27.29 (+6.22) \\
			 &\texttt{\$REPLACE\_\{t\}} & 2079 & 67.46 & 36.99 & 47.78 (+18.95) & 660 & 48.96 &28.64 &36.14 (+20.04) \\
			\hline
			
	         \multicolumn{1}{c}{\multirow{2}{*}{$D$(\texttt{DELETE})}} &\texttt{\$APPEND\_\{t\}}    & 1024 & 52.69 & 25.78 & 34.62 & 332 & 18.93 &13.86 &16.00 \\
		      &\texttt{\$REPLACE\_\{t\}}   & 1425 & 50.91 & 19.72 & 28.43 & 716 & 30.89 &13.55 &18.83 \\[0.2ex]
			\hdashline[2pt/2pt]	
			\multicolumn{1}{c}{\multirow{2}{*}{$D$(\texttt{DELETE\checkmark})}} &\texttt{\$APPEND\_\{t\}}  & 1024 & 57.14 & 27.73 & 37.34 (+2.72) & 332 & 22.77 &15.36 &18.35 (+2.35) \\
			 &\texttt{\$REPLACE\_\{t\}} & 1425 & 55.02 & 22.32 & 31.75 (+4.32) & 716 & 36.17 &16.62 &22.78 (+3.95) \\
			
			\hline
	        \multicolumn{1}{c}{\multirow{2}{*}{$D$(\texttt{REPLACE})}} &\texttt{\$APPEND\_\{t\}}    & 1762 & 52.76 & 29.34 & 37.71 & 443 & 23.92 &18.74 &21.01 \\
		     &\texttt{\$DELETE}    & 996 & 56.19 & 18.67 & 28.03 & 767 & 47.10 &15.91 &23.78 \\[0.2ex]
			\hdashline[2pt/2pt]	
			\multicolumn{1}{c}{\multirow{2}{*}{$D$(\texttt{REPLACE\checkmark})}} &\texttt{\$APPEND\_\{t\}} & 1762 & 68.05 & 49.21 & 57.11 (+19.40) & 443 & 41.97 &44.24 &43.08 (+22.07) \\
			 &\texttt{\$DELETE} & 996 & 69.33 & 34.04 & 45.66 (+17.63) & 767 & 61.08 &25.16 &35.64 (+11.86) \\
			\bottomrule
	
	\end{tabular}
	}
	\caption{Results of our quantitative experiments. $D$(\texttt{ACTION}) denotes a subset consisting of instances with \texttt{ACTION} label. $D$(\texttt{ACTION\checkmark}) denotes another version of $D$(\texttt{ACTION}), where corresponding errors have been manually corrected. }
	\label{tab:experiment_validation}
\end{table*}

\section{Effect of the Interdependence between Different Types of Corrections}\label{section:interdependence_of_correcting actions}

In this section, we conduct several groups of quantitative experiments to explore the interdependence between corrections.

We first train the GECToR model on \emph{Stage \uppercase\expandafter{\romannumeral2} Only} for efficiency. All training settings are the same to published parameters.\footnote{We use the codes of new version from \url{https://github.com/grammarly/gector/pull/120} after contacting authors.} Afterwards, we use the model to conduct corrections on the BEA-2019 (W\&I+LOCNESS) dev set and CoNLL-2014 test set \cite{ng_conll2014} and their variants with some errors corrected manually. For simplicity, we only consider the three most frequent editing-action labels: \texttt{\$APPEND\_\{t\}}, \texttt{\$DELETE} and \texttt{\$REPLACE\_\{t\}}.

Figure \ref{fig:control_expriment} shows the procedure of quantitative experiments. Specifically, we separate each raw erroneous sentence into five parts: correct words, erroneous words that can be corrected by \texttt{\$APPEND\_\{t\}}/\texttt{\$DELETE}/\texttt{\$REPLACE\_\{t\}}, and words with other types of errors. If we want to investigate the influence of \texttt{\$APPEND\_\{t\}}, we first select the data containing \texttt{\$APPEND\_\{t\}} labels and denote them as $D$(\texttt{APPEND}). Then we manually correct all the errors which should be corrected by \texttt{\$APPEND\_\{t\}} labels, obtaining the new subset $D$({\texttt{APPEND\checkmark}}). Afterwards, we use our model to correct erroneous sentences of subsets $D$(\texttt{APPEND}) and $D$({\texttt{APPEND\checkmark}}) for just one iteration, and finally we only evaluate and compare the model performance on the predictions of \texttt{\$DELETE} and \texttt{\$REPLACE\_\{t\}}. For example, by comparing the model performance with respect to the \texttt{\$DELETE} label, we can draw the conclusion that appending some words first could help the model to achieve better predictions on \texttt{\$DELETE} .

Likewise, we conduct experiments with respect to \texttt{\$DELETE} and \texttt{\$REPLACE\_\{t\}} labels. Besides, we evaluate the performance for each type of labels on the raw dataset without any constraints. Experimental results of the RoBERTa-based GECToR model \cite{Liu_arxiv2019} are listed in {Table \ref{tab:experiment_validation}}. 
We can observe that the consistent performance improvements occur on both the W\&I+LOCNESS dev set and the CoNLL-2014 test set, no matter which type of errors are corrected first. Moreover, it is surprising that if replacing words or appending words are conducted beforehand, the model performance is significantly improved on correcting other types of errors. Meanwhile, deleting words does not benefit others compared with other two kinds of corrections.

We also notice that the model improvements are positively associated with the number of manual corrections on the BEA-2019 dev set. However, the performance improvements on the CoNLL-2014 test set is not closely related to the number of manual corrections. Thus, we can conclude that the interdependence between different types of corrections indeed plays a more important role than the number of corrections on performance improvements.
Having witnessed these experimental results, we can arrive at the following two conclusions: 

\begin{itemize}
    \vspace{-5pt}
    \item GEC models can better deal with errors when some types of errors have been corrected.
    \vspace{-5pt}
    \item Corrections of appending words or replacing words help the model correct other types of errors more than deleting words.
    \vspace{-5pt}
    
\end{itemize}

Please note that we also conduct experiments using the XLNet-based GECToR model \cite{Yang_nips2019}. Similar trend can be observed from experimental results reported in Appendix \S\ref{subsec:appendix}.

\section{Our Approach}\label{sec:approach}

In this section, we introduce our proposed Type-Driven Multi-Turn Corrections (TMTC) approach in detail. As concluded above, correcting certain types of errors first benefits correcting others, thus, we decompose one-iteration corrections of each training instance into multi-turn corrections, so as to make the trained model to learn performing corrections progressively.

 \begin{figure}[ht]
  \centering
  \includegraphics[width=1\linewidth]{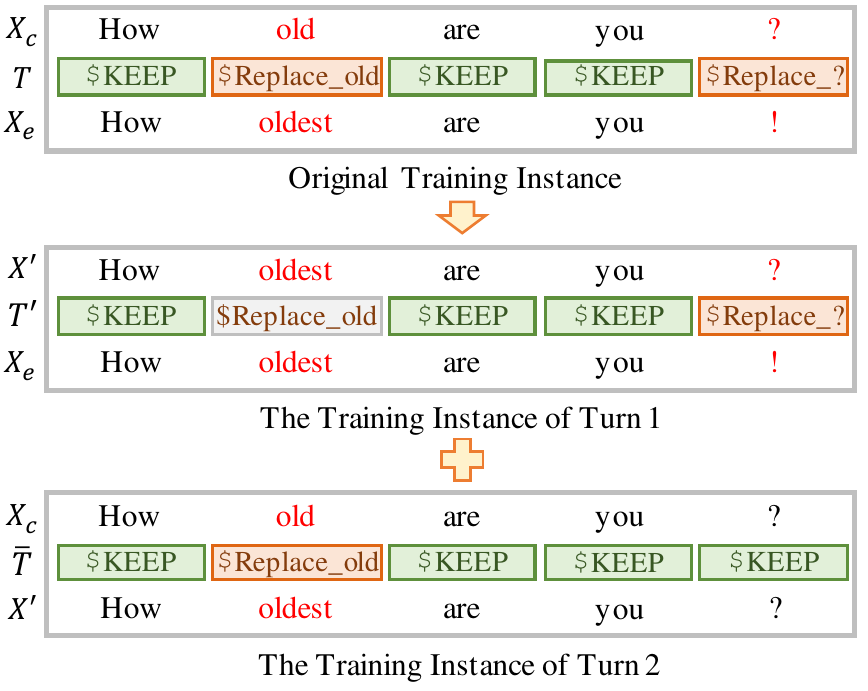}
  \vspace{-15pt}
  \caption{The procedure illustration of constructing additional training instances. Here, we construct an intermediate sentence $X'$, which is paired with the raw erroneous sentence $X_e$ and reference sentence $X_c$ to form two additional training instances $(X_{e},X')$ and $(X',X_{c})$, respectively. Red squares mean labels correcting errors, while green ones mean the labels to keeping the current word unchanged. Losses of gray squares will be omitted in the first turn.  }
  \label{fig:MTC_framework}

\end{figure}

The key step of our approach is to construct an intermediate sentence  for each training instance. Formally, each training instance is a sentence pair $(X_{e},X_{c})$ consisting an erroneous sentence $X_{e}$ and a reference sentence $X_{c}$. To construct its intermediate sentence $X'$, we randomly select partial grammatical errors and correct them manually while keeping others unchanged. Then, $X'$ is paired with $X_{e}$ and $X_{c}$ to generate two new pairs: $(X_{e},X')$ and $(X',X_{c})$, respectively. Figure \ref{fig:MTC_framework} illustrates an example of constructing two additional training instances from a sentence pair. In this example, for the 
erroneous sentence with two grammatical errors ``\textit{oldest}'' and ``\textit{!}'', we correct ``\textit{!}'' by ``\textit{?}'' manually to form the semi-corrected sentence ``\textit{How oldest are you ?}''.
It should be noted that our constructed training instances are derived from the original training corpus, and thus their grammatical errors are also human-making.

Based on the above findings mentioned in Section \S\ref{section:interdependence_of_correcting actions}, we apply our approach to design three training strategies: {\textbf{\texttt{APPEND}-first}}, \textbf{\texttt{DELETE}-first} and \textbf{\texttt{REPLACE}-first}. Here, the \texttt{ACTION}-first strategy means that the model is trained to learn \texttt{ACTION} corrections in the first turn and then the others in the second turn. For example, when using the \texttt{DELETE}-first strategy, we keep the errors with ``\texttt{\$DELETE}'' as target labels unchanged during the constructions of intermediate sentences.
Using additionally-constructed training instances involving these sentences, the trained model will be encouraged to focus on performing corrections first via \texttt{\$DELETE}. Table \ref{tab:con_num} lists the numbers of additionally-constructed training instances using these strategies. According to our findings concluded in Section \S\ref{section:interdependence_of_correcting actions}, the models trained using \texttt{APPEND}-first and \texttt{REPLACE}-first strategies should perform better.

\begin{table}[t] \footnotesize
	\centering
	\setlength{\tabcolsep}{1.5mm}{
		
		\begin{tabular}{lc} 
			\toprule
			 {Strategy}	& \#Additional Instance  \\
			\midrule
			\texttt{RANDOM}	 & 367,814  \\[0.2ex]
			\texttt{APPEND}-first & 311,348 \\[0.2ex]
			\texttt{DELETE}-first & 326,100 \\[0.2ex]
			\texttt{REPLACE}-first & 296,683 \\
			\bottomrule
	\end{tabular}
	}
	\caption{Numbers of additionally-constructed training instances. We also explore the training strategy that randomly corrects partial errors first. For convenience, we name this training strategy as \texttt{RANDOM}. }
	\label{tab:con_num}

\end{table}

Using our approach, 
we adopt different objectives to successively train our model. Specifically, we define the following training objectives $L_c^{(1)}$ and $L_c^{(2)}$ in the first and second turns, respectively:

\vspace{-10pt}
\begin{equation} \label{eq:new_turn_loss}
	\begin{aligned}
		L_c^{(1)}&=-\sum_{i=1}^{N} \mathbbm{1}(t'_i=t_i) \cdot \text{log} p(t'_i|X_e,\theta), \\ 
	\end{aligned}
\end{equation}
\vspace{-10pt}
\begin{equation} 
\begin{aligned}
	L_c^{(2)}&=-\sum_{i=1}^{\Bar{N}}\text{log} p(\Bar{t}_i|X',\theta),\\
\end{aligned}
\end{equation}where $\{t'_i\}_{i=1}^N$ and $\{\Bar{t}_i\}_{i=1}^{\Bar{N}}$ are the editing-action label sequences of additionally-constructed training instances $(X_{e},X')$ and $(X',X_{c})$ respectively.

Notably, there remain some grammatical errors within intermediate sentences
which not be learned by the model in the first turn. 
Therefore,
we omit the incorrect supervisal signals in the definition of $L_c^{(1)}$ via an
indicator function $\mathbbm{1}(*)$, which is used to shield the effect of incorrect losses.
However, because our additionally-constructed training instances contain less grammatical errors compared with original ones, which causes the trained model to correct less errors. 
To address this defect, we still use the original training instances to continuously train model in the third turn.

Formally, we finally,
we use all training instances to continuously train our model with the following objective $L'$=$L_c^{(1)}$+$L_c^{(2)}$+$L$.
Our experimental results presented in Section \S\ref{section:experiment} show that our additionally-constructed training instances and original ones are complementary to each other. 

\section{Experiment}\label{section:experiment}

\begin{table*}[ht]\small
	\centering
	\setlength{\tabcolsep}{1.5mm}{
		\begin{tabular}{cccccccccc} 
			\toprule
			\multicolumn{1}{c}{\multirow{2}{*}{Model}} & \multicolumn{1}{c}{\multirow{2}{*}{Pre-trained}} & &\multicolumn{3}{c}{BEA-2019 (dev)} & &\multicolumn{3}{c}{CoNLL-2014 (test)} \\
			  &  & & Prec. & Rec. & F$_{0.5}$ &   & Prec. & Rec. & F$_{0.5}$ \\
			\midrule
			\multicolumn{1}{c}{\multirow{2}{*}{GECToR\cite{omelianchuk_gector_ACL20}$^\dagger$}} 
			                        &RoBERTa &	 & 50.30 & 30.50 & 44.50 & & 67.50 & 38.30 & 58.60 \\[0.2ex]
			                        &XLNet  &	   & 47.10 & 34.20 & 43.80 & & 64.60 & 42.60 & 58.50 \\
			\hline
			\multicolumn{1}{c}{\multirow{2}{*}{GECToR}} 
			              &RoBERTa &     & 49.80 & 37.61 & 46.77 & & 66.56 & 45.08 & 60.77 \\[0.2ex]
			                &XLNet &     & 45.55 & 39.81 & 44.27 & & 64.04 & 48.67 & 60.24 \\
			\hline
			GECToR(\texttt{RANDOM})  &RoBERTa &  & 52.88 & 36.05 & 48.37 (+1.60) & & 69.54  & 44.32 & 62.43 (+1.66) \\[0.1ex]
			GECToR(\texttt{APPEND}-first) &RoBERTa & & 54.92 & 35.30 & \textbf{49.43} (+2.66) & & 70.73 & 43.88 & \textbf{63.01} (+2.24) \\[0.1ex]
			GECToR(\texttt{DELETE}-first) &RoBERTa & & 53.85 & 35.13 & 48.67 (+1.90) & & 70.57 &42.78 &62.45 (+1.68) \\[0.1ex]
			GECToR(\texttt{REPLACE}-first) &RoBERTa & & 54.78 & 34.82 & 49.14 (+2.37) & & 70.2 & 43.92 &62.70 (+1.93) \\[0.1ex]
			\hdashline[2pt/2pt]	
			GECToR(\texttt{RANDOM})  &XLNet &  & 49.74 & 38.47 & 46.99 (+2.72) & & 67.41 & 46.68 & 61.91 (+1.67) \\[0.1ex]
			GECToR(\texttt{APPEND}-first)  &XLNet & & 51.10 & 37.72 & 47.71 (+3.44) & & 67.74 & 46.39 & 62.03 (+1.79) \\[0.1ex]
			GECToR(\texttt{DELETE}-first)  &XLNet & & 50.48 & 37.49 & 47.21 (+2.97) & & 67.33 & 46.42 & 61.79 (+1.55) \\[0.1ex]
			GECToR(\texttt{REPLACE}-first) &XLNet & & 51.96 & 37.19 & \textbf{48.14} (+3.87) & & 69.36 & 46.30 & \textbf{63.08} (+2.84) \\
			\bottomrule
	\end{tabular}
	}
	\caption{Results of models in the dataset setting of Stage \uppercase\expandafter{\romannumeral2} Only. $\dagger$ indicates scores reported in previous papers.}
	\label{tab:stage_2_only}
\end{table*}

\begin{table*}[ht]\small
	\centering
	\setlength{\tabcolsep}{1.4mm}{
		\begin{tabular}{cccccccccc} 
			\toprule
			\multicolumn{1}{c}{\multirow{2}{*}{Model}} & \multicolumn{1}{c}{\multirow{2}{*}{Pre-trained}} & &\multicolumn{3}{c}{BEA-2019 (test)} & &\multicolumn{3}{c}{CoNLL-2014 (test)} \\
			 & & & Prec. & Rec. & F$_{0.5}$ &   & Prec. & Rec. & F$_{0.5}$ \\
			\midrule
			Dual-boost\cite{Ge_ACL18}$^\dagger$ & - &  & - & - & - &  & 64.47 & 30.48 & 52.72 \\[0.2ex]
			\hdashline[2pt/2pt]	
			\multicolumn{1}{c}{\multirow{2}{*}{GECToR\cite{omelianchuk_gector_ACL20}$^\dagger$}}
			                                             &RoBERTa &	 & 77.2 & 55.1 & 71.5 & & 72.1 & 42.0 & 63.0 \\[0.2ex]
			                                             &XLNet &	 & 79.2 & 53.9 & 72.4 & & 77.5 & 40.1 & 65.3 \\[0.2ex] 
            \hdashline[2pt/2pt]	
			\multicolumn{1}{c}{\multirow{2}{*}{GECToR(GST)\cite{Parnow_aclf21}$^\dagger$}}
			                                         &RoBERTa &       & 77.5 & 55.7 & 71.9 & & 74.1 & 42.2 & 64.4 \\[0.2ex]
			                                          &XLNet &        & 79.4 & 54.5 & 72.8 & & 78.4 & 39.9 & 65.7 \\[0.2ex]
			\hdashline[2pt/2pt]	
			SAD((12+2)\cite{Sun_ACL2021}$^\dagger$ &BART &	 & - & - & \textbf{72.9} & & 71.0 & 52.8 & \textbf{66.4} \\
			\hline
			\multicolumn{1}{c}{\multirow{2}{*}{GECToR}}  &RoBERTa &	 & 78.02 & 53.49 & 71.53 & & 72.93 & 40.02 & 63.11 \\[0.2ex]
			  &XLNet &	 & 80.23 & 51.76 & 72.36 & & 77.63 & 40.11 & 65.57 \\
			\hline
			GECToR(\texttt{RANDOM})  &RoBERTa &  & 79.85 & 51.53 & 71.94 (+ 0.41) & & 75.39  & 41.57 & 64.84 (+ 1.73) \\[0.1ex]
			GECToR(\texttt{APPEND}-first)  &RoBERTa & & 80.31 & 51.14 & 72.08 (+0.55) & & 76.77 & 40.95 & 65.34 (+2.23) \\[0.1ex]
			GECToR(\texttt{DELETE}-first)  &RoBERTa & & 79.39 & 52.25 & 71.92 (+0.39) & & 75.70 & 39.85 &64.16 (+1.05) \\[0.1ex]
			GECToR(\texttt{REPLACE}-first)  &RoBERTa & & 81.27 & 50.67 & \textbf{72.51} (+0.98) & & 77.36 & 40.35 & \textbf{65.37} (+ 2.26) \\[0.1ex]
			\hdashline[2pt/2pt]	
			GECToR(\texttt{RANDOM})  &XLNet &  & 81.14 & 50.83 & 72.49 (+0.13) & & 77.08 & 42.03 & 66.06 (+0.49) \\[0.1ex]
			GECToR(\texttt{APPEND}-first)  &XLNet & & 81.89 & 50.55 & 72.85 (+0.49) & & 78.18 & 42.67 & \textbf{67.02} (+1.45) \\[0.1ex]
			GECToR(\texttt{DELETE}-first)  &XLNet & & 82.35 & 49.52 & 72.71 (+0.35) & & 77.05 & 42.03 & 66.04 (+0.47) \\[0.1ex]
			GECToR(\texttt{REPLACE}-first) &XLNet & & 81.33 & 51.55 & \textbf{72.91} (+0.55) & & 77.83 & 41.82 & 66.40 (+0.83) \\
			\bottomrule
	\end{tabular}
	}
	\caption{Results of models at the dataset setting of Three Stages of Training.}
	\label{tab:three_stages}
\end{table*}
\subsection{Setup}

To ensure fair comparison, we train the GECToR models using the same training datasets and parameters as \cite{omelianchuk_gector_ACL20}, and then evaluate them on the BEA-2019 (W\&I+LOCNESS) dev, test set and the CoNLL-2014 test set. The details of the training data are listed in Table \ref{tab:datasets}. Following \cite{omelianchuk_gector_ACL20}, we conduct experiments in two dataset settings: Stage \uppercase\expandafter{\romannumeral2} Only and Three Stages of Training. Notably, in the latter setting, we only apply our approach at Stage \uppercase\expandafter{\romannumeral2} and Stage \uppercase\expandafter{\romannumeral3} for efficiency. Finally, we evaluate the model performance in terms of official ERRANT \cite{bryant_acl2017_errant} and $M^2$ scorer \cite{dahlmeier_naccl2012_m2scorer} respectively.

\subsection{Main Results and Analysis} \label{subsec:main}
\begin{small}
\begin{table*}[t]\small
	\centering
	\setlength{\tabcolsep}{0.7mm}{
		\begin{tabular}{ccccccccccc} 
			\toprule
			\multicolumn{1}{c}{\multirow{3}{*}{Dataset}} &\multicolumn{1}{c}{\multirow{3}{*}{Strategy}}  &\multicolumn{1}{c}{\multirow{3}{*}{Evaluation}}      &\multicolumn{8}{c}{RoBERTa} \\ 
			\cline{4-11}
			 & & &\multicolumn{4}{c}{BEA-2019 (dev)} &\multicolumn{4}{c}{CoNLL-2014 (test)} \\
			\cline{4-11}
			 & & & Num. & Prec. & Rec. & F$_1$ & Num. & Prec. & Rec. & F$_1$ \\
			\hline
		  \multicolumn{1}{c}{\multirow{2}{*}{$D$(\texttt{APPEND})}} & \multicolumn{1}{c}{\multirow{2}{*}{\texttt{APPEND}-first}} & \texttt{\$DELETE}    & 904 & 64.03 & 19.69 & 30.12 & 496 & 45.45 &9.07 &15.13 \\[0.2ex]
		   &  &\texttt{\$REPLACE\_\{t\}}   & 2079 & 52.54 & 19.38 & 28.32 & 660 & 34.83 &9.39 &14.80 \\[0.2ex]
			\hdashline[2pt/2pt]	
			\multicolumn{1}{c}{\multirow{2}{*}{$D$(\texttt{APPEND}\checkmark)}} & \multicolumn{1}{l}{\multirow{2}{*}{\texttt{APPEND}-first}} &\texttt{\$DELETE}  & 904 & 79.17 & 33.63 & 47.20 (+17.08) & 496 & 68.18 &18.15 &28.66 (+13.53) \\[0.2ex]
			 &  &\texttt{\$REPLACE\_\{t\}} & 2079 & 73.49 & 36.80 & 49.04 (+20.72) & 660 & 60.84 &28.48 &38.80 (+24.00) \\
			\hline
			
	      \multicolumn{1}{c}{\multirow{2}{*}{$D$(\texttt{DELETE})}} &\multicolumn{1}{c}{\multirow{2}{*}{\texttt{DELETE}-first}} &\texttt{\$APPEND\_\{t\}}    & 1024 & 54.31 & 22.75 & 32.07 & 332 & 24.53 &11.75 &15.89 \\[0.2ex]
		    & &\texttt{\$REPLACE\_\{t\}}   & 1425 & 52.75 & 18.88 & 27.80 & 716 & 35.19 &10.61 &16.31 \\[0.2ex]
			\hdashline[2pt/2pt]	
			\multicolumn{1}{c}{\multirow{2}{*}{$D$(\texttt{DELETE}\checkmark)}} &\multicolumn{1}{c}{\multirow{2}{*}{\texttt{DELETE}-first}} &\texttt{\$APPEND\_\{t\}}  & 1024 & 60.28 & 25.49 & 35.83 (+3.76) & 332 & 30.32 &14.16 &19.30 (+3.41) \\[0.2ex]
			& &\texttt{\$REPLACE\_\{t\}} & 1425 & 59.16 & 22.67 & 32.78 (+4.98) & 716 & 40.32 &13.97 &20.75 (+4.44) \\
			
			\hline
	      \multicolumn{1}{c}{\multirow{2}{*}{$D$(\texttt{REPLACE})}} & \multicolumn{1}{c}{\multirow{2}{*}{\texttt{REPLACE}-first}} &\texttt{\$APPEND\_\{t\}}    & 1762 & 55.32 & 27.13 & 36.41 & 443 & 28.74 &16.03 &20.58 \\[0.2ex]
		  & &\texttt{\$DELETE}    & 996 & 58.13 & 19.38 & 29.07 & 767 & 50.00 &11.34 &18.49 \\[0.2ex]
			\hdashline[2pt/2pt]	
			\multicolumn{1}{c}{\multirow{2}{*}{$D$(\texttt{REPLACE}\checkmark)}}&\multicolumn{1}{c}{\multirow{2}{*}{\texttt{REPLACE}-first}} &\texttt{\$APPEND\_\{t\}} & 1762 & 73.57 & 47.56 & 57.77 (+21.36) & 443 & 53.82 & 42.89 & 47.74 (+27.16) \\[0.2ex]
			 & &\texttt{DELETE} & 996 & 77.99 & 36.65 & 49.86 (+20.79) & 767 & 71.75 & 25.16 & 37.26 (+18.77) \\
			\bottomrule
	
	\end{tabular}
	}
	\caption{Results of our quantitative experiments using models enhanced by our approach. Three groups of experiments are conducted on the same data subset as Table \ref{tab:experiment_validation}.}
	\label{tab:experiment_validation_MTC}
\end{table*}        
\end{small}

\textbf{Stage \uppercase\expandafter{\romannumeral2} Only.} In this setting, we compare the performance of GECToR with or without applying our approach\footnote{Please note that previous studies do not provide the performance of other baselines under the setting of Stage \uppercase\expandafter{\romannumeral2} Only.}. 

Results are presented on Table \ref{tab:stage_2_only}. Notably, the results are consistent with our findings in Section \S\ref{section:interdependence_of_correcting actions}. That is, since correcting some types of errors benefit the corrections of other errors, all models trained with our approach significantly perform better than their corresponding baselines. Moreover, the GECToR models trained by the \texttt{APPEND}-first or \texttt{REPLACE}-first strategies are superior to models trained by \texttt{DELETE}-first or \texttt{RANDOM}, echoing the conclusions mentioned in Section \S\ref{section:interdependence_of_correcting actions}.

\textbf{Three Stages of Training.}  In this setting, we compare our enhanced models with more baselines under the setting of the single model, including the most related work, Dual-boost \cite{Ge_ACL18}, GECToR(GST) \cite{Parnow_aclf21} and the current best NMT-based model SAD(12+2) \cite{Sun_ACL2021}. 

As reported in Table \ref{tab:three_stages}, we obtain the similar results to Stage \uppercase\expandafter{\romannumeral2} Only. Our approach promotes models to obtain desirable improvements, where the \texttt{APPEND}-first and \texttt{REPLACE}-first strategies perform better. Overall, the GECToR models trained by our approach are comparable or even better than SAD(12+2). Particularly, when ensembling our enhanced models with competitive GEC models, we obtain 77.93 $F_{0.5}$, achieving SOTA score on the BEA-2019 test set.

Moreover, we find that our approach allows the trained models to correct more cautiously. That is, the trained models tend to perform less but more precise corrections, compared with the basic GECToR models. One of underlying reasons is that our additionally-constructed training instances contain more \texttt{\$KEEP} labels especially in the second turn, which makes the label predictions of the model biased.

\subsection{Ablation Study} \label{subsec:ablation}

Then, we conduct more experiments to investigate the effectiveness of various details on our proposed approach.

\begin{small}
\begin{table}[t]\footnotesize
	\centering
	\setlength{\tabcolsep}{0.8mm}{
		\begin{tabular}{lcccccccc} 
			\toprule
			\multicolumn{1}{c}{\multirow{2}{*}{Model}} &\multicolumn{3}{c}{BEA-2019 (dev)} &  &\multicolumn{3}{c}{CoNLL-2014 (test)} \\
			   & Prec. & Rec. & F$_{0.5}$ &   & Prec. & Rec. & F$_{0.5}$ \\
			\midrule
			GECToR & 49.80 & 37.61 & 46.77 & & 66.56 & 45.08 & 60.77 \\
			\quad w/ TMTC  & 54.92 & 35.30 & 49.43 & & 70.73 & 43.88 &63.01 \\
            \quad w/o turn 1  & 51.29 & 37.01 & 47.03 & & 68.99 & 45.45 & 62.51 \\
            \quad w/o turn 2  & 50.43  & 37.3  & 47.12 & & 66.94 & 44.60 & 61.31 \\
            \quad w/o orignal  & 55.21 & 32.5 & 48.44 & & 71.22 & 41.55 & 62.32 \\
            \quad mix data  & 53.04 & 31.00 & 46.44 & & 71.31 & 40.59 & 61.84  \\
            \quad w/o $\mathbbm{1}(*)$  & 53.23 & 33.49 & 47.62 & & 71.31 & 42.16 &62.64 \\
			\bottomrule
	\end{tabular}
	}
	\caption{Ablation study. Our model is based on RoBERTa and trained using \texttt{APPEND}-first. The $\mathbbm{1}(*)$ is the indicator function mentioned in Equation \ref{eq:new_turn_loss}.}
	\label{tab:ablation}
\end{table}
\end{small}

All experimental results are provided in Table \ref{tab:ablation}. 
Results of lines 3-5 (``w/o turn 1'', ``w/o turn 2'', ``w/o original'') demonstrate that our additionally-constructed training instances are complementary to original ones.
In addition, we also directly mix the additionally-constructed training instances and the original ones to train a GECToR model. However, such a training strategy does not promote the model to learn much better, showing the advantage of gradual learning error corrections.
Finally, as mentioned in Section \S\ref{sec:approach}, some grammatical errors should not be learned within intermediate sentence. Here, we also report the performance of the GECToR model without omitting incorrect supervisal signals. As shown in the line 7 (``w/o $\mathbbm{1}(*)$'') of Table \ref{tab:ablation}, the lower recall values indicate these incorrect \texttt{\$KEEP} labels make the model to infer more conservatively.

\begin{figure}[t]
  \centering
  \includegraphics[width=0.95\linewidth]{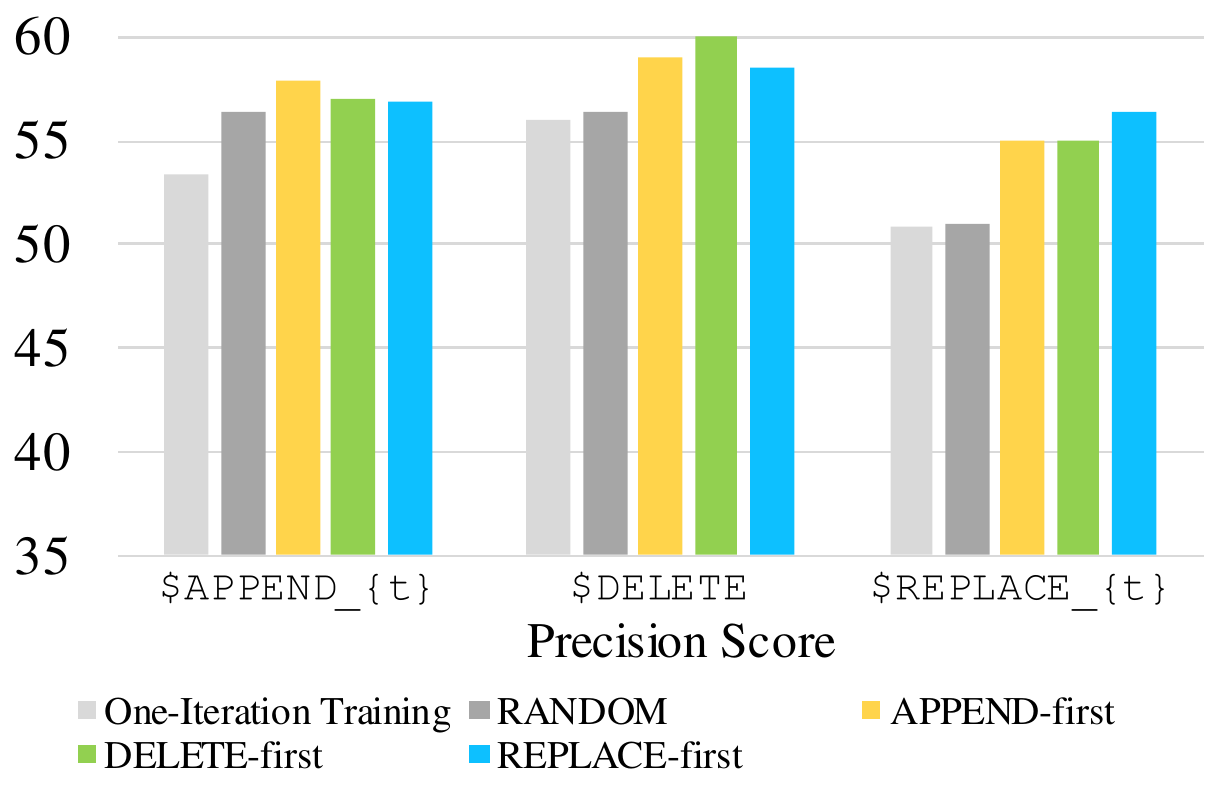}
  \caption{Label predictions of the RoBERTa-based model on the BEA-2019 dev set in the first iteration of prediction.}
  \label{fig:tendency}
\end{figure}

\subsection{Analysis}

\begin{table*}[t]\footnotesize
	\centering
	\setlength{\tabcolsep}{2mm}{
		\begin{tabular}{lcccccccc} 
			\toprule
			\multicolumn{1}{c}{\multirow{2}{*}{Model}} &\multicolumn{3}{c}{BEA-2019 (dev)} &  &\multicolumn{3}{c}{CoNLL-2014 (test)} \\
			   & Prec. & Rec. & F$_{0.5}$ &   & Prec. & Rec. & F$_{0.5}$ \\
			\midrule
			\multicolumn{1}{c}{GECToR} & 49.80 & 37.61 & 46.77 & & 66.56 & 45.08 & 60.77 \\[0.1ex]
            GECToR(\texttt{APP+REP+DEL})  & 59.26 & 31.70 & 50.48 & & 74.08 & 40.37 & 63.48 \\[0.1ex]
            GECToR(\texttt{APP+DEL+REP})  & 58.38 & 32.06 & 50.15 & & 73.26 & 40.89 & 63.24  \\[0.1ex]
            GECToR(\texttt{REP+APP+DEL})  & 57.75 & 30.95 & 49.23 & & 74.36 & 39.19 & 63.05 \\[0.1ex]
            GECToR(\texttt{REP+DEL+APP})  & 57.72 & 31.44 & 49.66 & & 73.86 & 39.87 & 62.88  \\[0.1ex]
            GECToR(\texttt{DEL+APP+REP})  & 59.13 & 31.52 & 50.04 & & 74.28 & 39.61 & 63.15 \\[0.1ex]
            GECToR(\texttt{DEL+REP+APP})  & 58.51 & 31.83 & 50.18 & & 73.34 & 40.55 & 63.06 \\
			\bottomrule
	\end{tabular}
	}
	\caption{Results of more fine-grained strategies. We conduct experiments by the model trained at Stage \uppercase\expandafter{\romannumeral2} Only based on RoBERTa.}
	\label{tab:more_phases}

\end{table*}

\textbf{Correction Trend.} Here, we use the models trained under different strategies to not only evaluate the one-iteration performance with respect to our investigated three types of labels, but also conduct quantitative experiments again. By doing so, we can investigate if our approach indeed guides the model to correct some types of error first.

As shown in Figure \ref{fig:tendency}, we find our strategies indeed guide model to correct corresponding errors more precisely in the first iteration. Meanwhile, the less but more precise predictions occur again with respect to corresponding labels. For example, when only considering the model performance with respect to \texttt{\$APPEND\_\{t\}}, we observe that the model trained by \texttt{APPEND}-first obtains the highest precision score.

More importantly, back to Table \ref{tab:experiment_validation_MTC}, the phenomenon that correcting some types of errors benefits the others is highlighted. It indicates that our approach indeed allows the trained model to capture the interdependence between different types of corrections. 

\begin{figure}[t]
  \centering
  \includegraphics[width=0.9\linewidth]{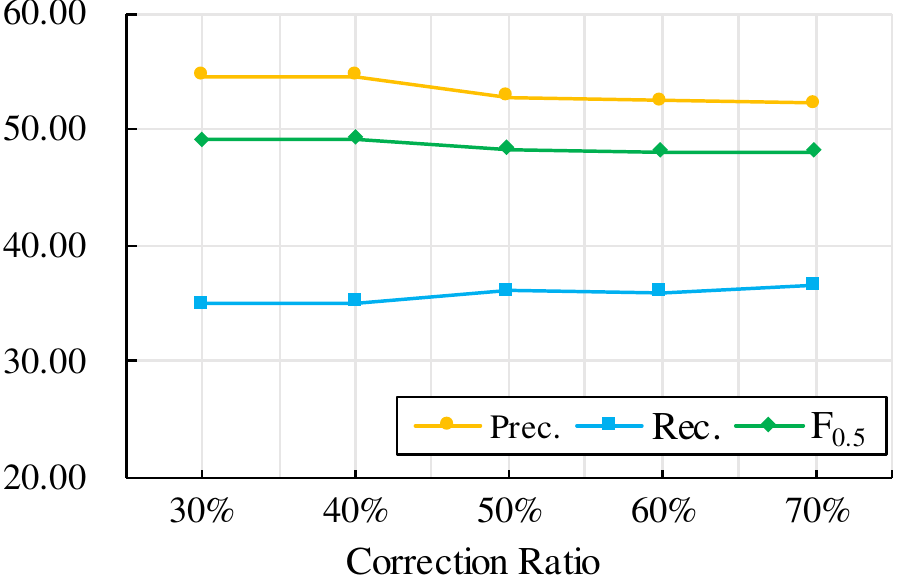}
  \vspace{-4pt}
  \caption{The precision, recall and F$_{0.5}$ values with respect to different correction ratios. }
  \label{fig:intermediate_sentence}
\end{figure}

\textbf{Effect of Correction Ratio.} As described in Section \S\ref{sec:approach}, the correction ratio is an important hyper-parameter that determines the numbers of manual corrections. Thus, we try different correction ratio values to investigate its effect on our approach. Figure \ref{fig:intermediate_sentence} shows the performance of the trained model with varying correction ratios. Apparently, with the correction ratio increasing, the precision score drops and recall score rises.  By contrast, the overall F$_{0.5}$ scores are always steady.

\begin{figure}[ht]
  \centering
  \includegraphics[width=0.86\linewidth]{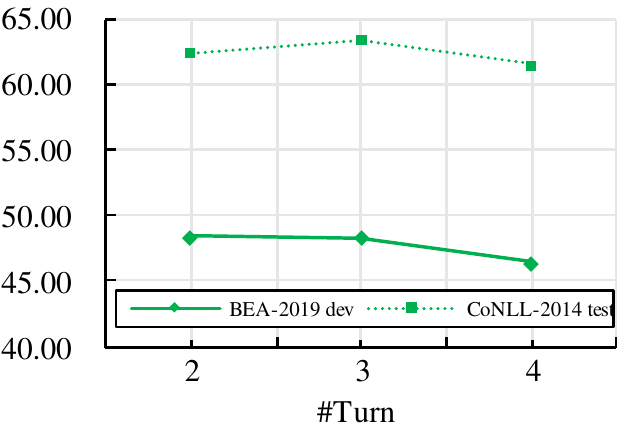}
  \vspace{-6pt}
  \caption{The F$_{0.5}$ scores of GECToR(\texttt{RANDOM}) with more turns of corrections.}
  \label{fig:mtc_more_turn}
  \vspace{-7pt}
\end{figure}

\textbf{Effect of More Turns of Corrections.} The above experimental results show that decomposing the conventional one-iteration training of into the two-turn training is useful to improve model training. A natural problem arises: can the trained model be further improved if we use more turns of training?

To answer this question, we use the model trained by the \texttt{RANDOM} strategy to conduct experiments. Specifically, we decompose the one-iteration corrections into $K$ turns of corrections, where we construct intermediate sentence by accumulatively correct $\frac{1}{K}$ errors during each turn of corrections. From Figure \ref{fig:mtc_more_turn}, we can observe that more turns of corrections do not benefit our models over two-turn corrections under the \texttt{RANDOM} strategy while with more training cost.

Also, we conduct experiments using more fine-grained strategies. For example, we can design a training strategy: after learning corrections of \texttt{\$APPEND\_\{t\}}, the model learns to correct errors of \texttt{\$REPLACE\_\{t\}} and then to correct others. For convenience, we name this strategy as \texttt{APP+REP+DEL}, where APP, REP and DEL are abbreviations of \texttt{\$APPEND\_\{t\}}, \texttt{\$REPLACE\_\{t\}} and \texttt{\$DELETE}, respectively. As illustrated in Table \ref{tab:more_phases}, all models trained by our approach obtain slightly better performance when introducing more iterations of corrections. However, they require almost 1.5x training time compared with our standard TMTC approach.

\section{Conclusion}

In this paper, we have firstly conducted quantitative experiments to explore the interdependence between different types of corrections, with the finding that performing some types of corrections such as appending or replacing words first help models to correct other errors. Futhermore, we propose a Type-Driven Multi-Turn Corrections (TMTC) approach for GEC, which allows the trained model to be not only explicitly aware of the progressive corrections, but also exploit the interdependence between different types of corrections. Extensive experiments show that our enhanced model is able to obtain comparable or better performance compared with the SOTA GEC model.

In the future, we plan to apply bidirectional decoding \cite{zhang_aaai2018,su_ai2019,zhang_TASLP19} to further improve our approach. Besides, inspired by the recent syntax-aware research \cite{li_aclfinding2021}, we will explore the interdependence between corrections from other perspectives for GEC such as syntax.

\section*{Acknowledgment}
The project was supported by  
National Key Research and Development Program of China (No. 2020AAA0108004), 
National Natural Science Foundation of China (No. 61672440), 
Natural Science Foundation of Fujian Province of China (No. 2020J06001),
and Youth Innovation Fund of Xiamen (No. 3502Z20206059).
We also thank the reviewers for their insightful comments.

\bibliography{anthology,custom}
\bibliographystyle{acl_natbib}

\clearpage
\appendix
\section{Appendix}
\begin{table*}[h]\small
	\centering
	\setlength{\tabcolsep}{1.5mm}{
		\begin{tabular}{cccccccccc} 
			\toprule
			     &    &\multicolumn{8}{c}{XLNet} \\ 
			\cline{3-10}
			{Dataset} & Evaluation &\multicolumn{4}{c}{BEA-2019 (dev)} &\multicolumn{4}{c}{CoNLL-2014 (test)} \\
			\cline{3-10}
			& & Num. & Prec. & Rec. & F1 & Num. & Prec. & Rec. & F1 \\
			\hline
		   &\texttt{\$APPEND\_\{t\}} & 2609 & 50.61 & 38.06 & 43.45 & 621 & 24.4 & 26.09 & 25.21 \\
		  Original Dataset &\texttt{\$DELETE} & 1403 & 52.79 & 25.66 & 34.53 & 1115 & 49.65 & 19.01 & 27.50 \\
		   &\texttt{\$REPLACE\_\{t\}} & 3495 & 49.10 & 24.12 & 32.35 & 1398 & 37.06 & 20.89 & 26.72 \\
			\hline
		  \multicolumn{1}{c}{\multirow{2}{*}{$D$(\texttt{APPEND})}} &\texttt{\$DELETE}    & 904 & 61.89 & 21.02 & 31.38 & 496 & 46.34 & 15.32 & 23.03 \\
		   &\texttt{\$REPLACE\_\{t\}}   & 2079 & 50.65 & 20.68 & 29.37 & 660 & 32.30 & 14.24 & 19.77 \\
			\hdashline[2pt/2pt]	
			\multicolumn{1}{c}{\multirow{2}{*}{$D$(\texttt{APPEND\checkmark})}} &\texttt{\$DELETE}  & 904 & 72.66 & 30.86 & 43.32 (+11.94) & 496 & 68.18 & 18.15 & 28.66 (+5.63) \\
			 &\texttt{\$REPLACE\_\{t\}} & 2079 & 67.13 & 36.84 & 47.58 (+18.21) & 660 & 60.84 & 28.48 & 38.80 (+19.03) \\
			\hline
			
	      \multicolumn{1}{c}{\multirow{2}{*}{$D$(\texttt{DELETE})}} &\texttt{\$APPEND\_\{t\}}    & 1024 & 50.27 & 27.44 & 35.50 & 332 & 18.09 & 16.57 & 17.30 \\
		   &\texttt{\$REPLACE\_\{t\}}   & 1425 & 49.57 & 20.00 & 28.50 & 716 & 28.12 & 14.80 & 19.40 \\
			\hdashline[2pt/2pt]	
			\multicolumn{1}{c}{\multirow{2}{*}{$D$(\texttt{DELETE\checkmark})}} &\texttt{\$APPEND\_\{t\}}  & 1024 & 54.91 & 28.42 & 37.45 (+1.95) & 332 & 30.32 & 14.16 & 19.30 (+2.00) \\
			 &\texttt{\$REPLACE\_\{t\}} & 1425 & 51.40 & 21.89 & 30.71 (+2.21) & 716 & 40.32 & 13.97 & 20.75 (+1.35) \\
			
			\hline
	      \multicolumn{1}{c}{\multirow{2}{*}{$D$(\texttt{REPLACE})}} &\texttt{\$APPEND\_\{t\}}    & 1762 & 55.32 & 31.38 & 38.85 & 443 & 20.28 & 19.86 & 20.07 \\
		   &\texttt{\$DELETE}    & 996 & 56.37 & 20.88 & 30.48 & 767 & 45.16 & 16.43 & 24.09 \\
			\hdashline[2pt/2pt]	
			\multicolumn{1}{c}{\multirow{2}{*}{$D$(\texttt{REPLACE\checkmark})}} &\texttt{\$APPEND\_\{t\}} & 1762 & 65.47 & 50.91 & 57.28 (+18.43) & 443 & 53.82 & 42.89 & 47.74 (+27.67) \\
			&\texttt{\$DELETE} & 996 & 70.89 & 35.94 & 47.70 (+17.22) & 767 & 71.75 & 25.16 & 37.26 (+16.51) \\
			\bottomrule
	
	\end{tabular}
	}
	\caption{Results of our control expriment. Four groups of results are obtained by the same re-implemented GECToR model. }
	\label{tab:experiment_validation_XLnet}
\end{table*}

\subsection{Quantitative Experiments on XLNet} \label{subsec:appendix}
We also conduct quantitative experiments described in Section \S\ref{section:interdependence_of_correcting actions} using model trained based on XLNet. The overall results are closely similar to Table \ref{tab:experiment_validation}, which indicates that our findings and conclusions are not specific to a certain model or a certain dataset, but common among realistic human-making datasets.

\subsection{Evaluation on JFLEG}
Suggested by reviewers, we evaluate our approach on the JFLEG \cite{napoles-sakaguchi-tetreault:2017:EACLshort} dataset which focus on fluency. As shown in Table \ref{tab:stage_2_only_jfleg} and Table \ref{tab:three_stages_jfleg}, models trained by our approach obtain higher GLEU \cite{heilman-EtAl:2014:P14-2} compared with baselines, which demonstrate the effectiveness of decomposing one-iteration correction into multiple turns. However, editing-action based interdependence seems not very beneficial from the view of fluency.

\begin{table*}[t]\small
	\centering
	\setlength{\tabcolsep}{0.5mm}{
		\begin{tabular}{cccccccccccc} 
			\toprule
			\multicolumn{1}{c}{\multirow{2}{*}{Model}} &\multicolumn{1}{c}{\multirow{2}{*}{Pre-trained}} & &\multicolumn{3}{c}{BEA-2019 (dev)} & &\multicolumn{3}{c}{CoNLL-2014 (test)} & &{JFLEG (test)} \\
			  &  & & Prec. & Rec. & F$_{0.5}$ &   & Prec. & Rec. & F$_{0.5}$ & &GLEU\\
			\midrule
			\multicolumn{1}{c}{\multirow{2}{*}{GECToR\cite{omelianchuk_gector_ACL20}$^\dagger$}} &RoBERTa &
			               & 50.30 & 30.50 & 44.50 & & 67.50 & 38.30 & 58.60 & & -\\
			 &XLNet  &	   & 47.10 & 34.20 & 43.80 & & 64.60 & 42.60 & 58.50 & & - \\ 
			\hline
			\multicolumn{1}{c}{\multirow{2}{*}{GECToR}} &RoBERTa &   
			              & 49.80 & 37.61 & 46.77 & & 66.56 & 45.08 & 60.77 & & 42.75\\
			 &XLNet &     & 45.55 & 39.81 & 44.27 & & 64.04 & 48.67 & 60.24 & & 42.90 \\
			\hline
			GECToR(\texttt{RANDOM})  &Roberta &  & 52.88 & 36.05 & 48.37 (+1.60) & & 69.54  & 44.32 & 62.43 (+1.66) & & \textbf{56.64} \\
			GECToR(\texttt{APPEND}-first) &Roberta & & 54.92 & 35.30 & \textbf{49.43} (+2.66) & & 70.73 & 43.88 & \textbf{63.01} (+2.24) & & 56.61 \\
			GECToR(\texttt{DELETE}-first) &Roberta & & 53.85 & 35.13 & 48.67 (+1.90) & & 70.57 &42.78 &62.45 (+1.68) & & 56.48 \\
			GECToR(\texttt{REPLACE}-first) &Roberta & & 54.78 & 34.82 & 49.14 (+2.37) & & 70.2 & 43.92 &62.70 (+1.93) & & 55.97 \\
			\hdashline[2pt/2pt]	
			GECToR(\texttt{RANDOM})  &XLNet &  & 49.74 & 38.47 & 46.99 (+2.72) & & 67.41 & 46.68 & 61.91 (+1.67) & & 56.84 \\
			GECToR(\texttt{APPEND}-first)  &XLNet & & 51.10 & 37.72 & 47.71 (+3.44) & & 67.74 & 46.39 & 62.03 (+1.79) & & \textbf{57.15} \\
			GECToR(\texttt{DELETE}-first)  &XLNet & & 50.48 & 37.49 & 47.21 (+2.97) & & 67.33 & 46.42 & 61.79 (+1.55) & & 56.60 \\
			GECToR(\texttt{REPLACE}-first) &XLNet & & 51.96 & 37.19 & \textbf{48.14} (+3.87) & & 69.36 & 46.30 & \textbf{63.08} (+2.84) & & 56.73 \\
			
			\bottomrule
	\end{tabular}
	}
	\caption{Results of models under the settings of Stage \uppercase\expandafter{\romannumeral2} Only. $\dagger$ indicates scores reported in previous papers.}
	\label{tab:stage_2_only_jfleg}
\end{table*}

\begin{table*}[t]\small
	\centering
	\setlength{\tabcolsep}{0.5mm}{
		\begin{tabular}{cccccccccccc} 
			\toprule
			\multicolumn{1}{c}{\multirow{2}{*}{Model}} & \multicolumn{1}{c}{\multirow{2}{*}{Pre-trained}} & &\multicolumn{3}{c}{BEA-2019 (test)} & &\multicolumn{3}{c}{CoNLL-2014 (test)} & &{JFLEG (test)} \\
			 & & & Prec. & Rec. & F$_{0.5}$ &   & Prec. & Rec. & F$_{0.5}$ & & GLEU \\
			\midrule
			Dual-boost\cite{Ge_ACL18}$^\dagger$ & &  & - & - & - &  & 64.47 & 30.48 & 52.72 \\
			\hdashline[2pt/2pt]	
			\multicolumn{1}{c}{\multirow{2}{*}{GECToR\cite{omelianchuk_gector_ACL20}$^\dagger$}}
			                                              &RoBERTa &	 & 77.2 & 55.1 & 71.5 & & 72.1 & 42.0 & 63.0 & &- \\
			 &XLNet &	 & 79.2 & 53.9 & 72.4 & & 77.5 & 40.1 & 65.3 & &- \\ 
            \hdashline[2pt/2pt]	
			\multicolumn{1}{c}{\multirow{2}{*}{GECToR(GST)\cite{Parnow_aclf21}$^\dagger$}} &RoBERTa &       & 77.5 & 55.7 & 71.9 & & 74.1 & 42.2 & 64.4 & &- \\
			 &XLNet &        & 79.4 & 54.5 & 72.8 & & 78.4 & 39.9 & 65.7 & &- \\
			\hdashline[2pt/2pt]	
			SAD((12+2)\cite{Sun_ACL2021}$^\dagger$ &BART &	 & - & - & \textbf{72.9} & & 71.0 & 52.8 & \textbf{66.4} & & - \\
			\hline
			\multicolumn{1}{c}{\multirow{2}{*}{GECToR}}  &RoBERTa &	 & 78.02 & 53.49 & 71.53 & & 72.93 & 40.02 & 63.11 & & 42.96 \\
			  &XLNet &	 & 80.23 & 51.76 & 72.36 & & 77.63 & 40.11 & 65.57 & & 43.11 \\ 
			\hline
			GECToR(\texttt{RANDOM})  &Roberta &  & 79.85 & 51.53 & 71.94 (+ 0.41) & & 75.39  & 41.57 & 64.84 (+ 1.73) & & \textbf{59.05} \\
			GECToR(\texttt{APPEND}-first)  &Roberta & & 80.31 & 51.14 & 72.08 (+0.55) & & 76.77 & 40.95 & 65.34 (+2.23) & & 58.88  \\
			GECToR(\texttt{DELETE}-first)  &Roberta & & 79.39 & 52.25 & 71.92 (+0.39) & & 75.70 & 39.85 &64.16 (+1.05) & & 58.94 \\
			GECToR(\texttt{REPLACE}-first)  &Roberta & & 81.27 & 50.67 & \textbf{72.51} (+0.98) & & 77.36 & 40.35 & \textbf{65.37} (+ 2.26) & & 59.03 \\
			\hdashline[2pt/2pt]	
			GECToR(\texttt{RANDOM})  &XLNet &  & 81.14 & 50.83 & 72.49 (+0.13) & & 77.08 & 42.03 & 66.06 (+0.49) & & \textbf{58.73} \\
			GECToR(\texttt{APPEND}-first)  &XLNet & & 81.89 & 50.55 & 72.85 (+0.49) & & 78.18 & 42.67 & \textbf{67.02} (+1.45) & & 58.64 \\
			GECToR(\texttt{DELETE}-first)  &XLNet & & 82.35 & 49.52 & 72.71 (+0.35) & & 77.05 & 42.03 & 66.04 (+0.47) & & 58.45 \\
			GECToR(\texttt{REPLACE}-first) &XLNet & & 81.33 & 51.55 & \textbf{72.91} (+0.55) & & 77.83 & 41.82 & 66.40 (+0.83) & & 58.42 \\

			\bottomrule
	\end{tabular}
	}
	\caption{Results of models under the settings of Three Stages of Training.}
	\label{tab:three_stages_jfleg}
    \vspace{-7pt}
\end{table*}

\end{document}